\pdfoutput=1
\documentclass{article} 
\usepackage{iclr2016_conference,times}
\usepackage{hyperref}
\usepackage{url}
\usepackage{amsmath}
\usepackage{amsfonts}
\usepackage[pdftex]{graphicx}
\usepackage{subfig}
\usepackage{multirow}

\title{Identifying the Absorption Bump with Deep Learning}

\author{
Min Li, Sudeep Gaddam, Xiaolin Li \\
Department of Electrical and Computer Engineering\\
University of Florida\\
Gainesville, FL 32611, USA \\
\texttt{\{minli, sudeepgaddam\}@ufl.edu, andyli@ece.ufl.edu} \\
\And
Yinan Zhao, Jingzhe Ma, Jian Ge \\
Department of Astronomy \\
University of Florida \\
Gainesville, FL 32611, USA \\
\texttt{\{yinanzhao, jingzhema\}@ufl.edu, jge@astro.ufl.edu} \\
}

\newcommand{\angstrom}{\text{\normalfont\AA \ }}
\makeatletter
\newcommand{\ion}[2]{#1\ \textsc{\rmfamily\@roman{#2}}\relax}
\makeatother

\begin{document}

\maketitle

\begin{abstract}

The pervasive interstellar dust grains provide significant insights to understand the formation and evolution of the stars, planetary systems, and the galaxies, and may harbor the building blocks of life. One of the most effective way to analyze the dust is via their interaction with the light from background sources. The observed extinction curves and spectral features carry the size and composition information of dust. The broad absorption bump at 2175 \angstrom is the most prominent feature in the extinction curves. Traditionally, statistical methods are applied to detect the existence of the absorption bump. These methods require heavy preprocessing and the co-existence of other reference features to alleviate the influence from the noises. In this paper, we apply Deep Learning techniques to detect the broad absorption bump. We demonstrate the key steps for training the selected models and their results. The success of Deep Learning based method inspires us to generalize a common methodology for broader science discovery problems. We present our on-going work to build the DeepDis system for such kind of applications. 

\end{abstract}

\section{Introduction}
\label{sec:intro}

The pervasive interstellar dust contains footprints of how the cosmos has been evolving. Dust grains come from condensation of heavy elements produced by stars. They are considered the primary reaction sites for molecules to form and essentially the source for all the $H_2$ in the interstellar medium (\cite{gould1963interstellar, hollenbach1971surface}). Dust grains also play an important role in controlling the gas-phase metal abundances and the thermodynamic evolution of the interstellar medium. Besides, coagulation of interstellar dust grains in a protostellar disk, along with their catalyzed complex organic molecules, eventually leads to planets. Dust is thought to harbor the secrets of planetary formation, and even life.

Research on dust and gas is conducted by analyzing the reddening and extinction effects on the spectra of background light sources and re-radiated emission. In the Local Universe, it is feasible to compare stellar spectra and explore the extinction curves to investigate the dust grains. However, more luminous background light sources, such as quasars, are necessary to probe the dust and gas ingredients in the galaxies at larger distances.

Among the research topics on cosmic dust grains, the 2175 \angstrom broad absorption bump \footnote{We use the 2175 \angstrom broad absorption bump and absorption bump for short in the rest of this paper.} stands out with significant values. Although the precise characteristics have not yet been established, the 2175 \angstrom broad absorption bump is believed to be tightly bounded with some types of aromatic carbonaceous materials (\cite{draine2003interstellar, li2003dust}). The most promising carriers of the bump are the Polycyclic Aromatic Hydrocarbon (PAH) molecules. PAHs are recognized as the most abundant organic molecules in our Milk Way and other neighboring galaxies (\cite{peeters2003unidentified}), and are believed to be the building blocks of organic life (\cite{bernstein2002side, caro2002amino}).

The traditional way of discovering the absorption bump mainly depends on statistic techniques (\cite{jiang2011toward}). It generally involves three steps. First, a composite quasar spectrum is constructed using median combining (\cite{jiang2010dusty, jiang2010high}). The composite spectrum is then reddened and used to fit every spectrum from the candidate absorption bump spectra with a parameterized absorption profile. Finally, by applying constrains on peak position, bump width and bump height, a large portion of spectra without bump features are filtered out. In order to further exclude the false positives caused by some noises, a simulation technique (\cite{jiang2010dusty, jiang2010high}) is applied to determine the detection significance. Among the steps, curve fitting occupies considerable amount of time. In order to convolve the absorption bump with the composite spectrum, extra information is needed to determine the absorption redshift. This significantly constrains the available observations due to the lack of this extra information in some cases. While the composite quasar spectrum could potentially be reused for other related computations, the fitting procedure evolves iterative error minimization to get the best bump profile using curve fitting method, and is required for each and every new observation. The fitting is restricted to relatively smooth continuum of the quasar spectrum. However, the complicated types of emission and absorption in the candidate spectra could potentially disturb the fitting process, reducing the effectiveness of the method. All these issues are tightly coupled with the method itself, and cannot be easily resolved.

In this paper, we propose to apply a Deep Learning based method to detect the absorption bump and try to alleviate the aforementioned issues. Deep Learning is recently recognized with its ability to automatically extract high level features and accurately recognize or classify the target objects (\cite{hinton2006reducing, lecun2015deep}). Deep Learning is most effective when the model is trained with sufficient data and the features are complex. For the absorption bump application, a normal observed spectrum consists of over 4,000 raw features. Each of them is a flux value at a certain wavelength. Due to the complex cosmos environment, the relation space among the features could be complicated. We also observe that in the traditional method, the curving fitting process is actually generating various spectra with bumps. These two facts make Deep Learning model a natural fit for such an application. 

For the rest of this paper, we present the details of our Deep Learning based method for the absorption bump detection. Specifically, we first give more background information about the problem and the traditional method. We then show how we generate the raw training and testing data sets. Next, we present how we transform the raw data into different formats for models including fully connected neural networks and two convolutional neural networks. The details and the results are also reported. Finally, we present a generalization from our methodology to simplify and improve the efficiency for similar astronomy problems, and also resembling problems in other science fields. We discuss the workflow and the design details of our ongoing work, the DeepDis system, which aims to provide such services.

\section{Traditional method for discovering absorption bump}
\label{sec:trad-method}

Previous research (\cite{fitzpatrick1990analysis}) showed that the extinction curve and the variation of the intrinsic spectral slopes of a quasar can be represented by a parameterized linear component. A Drude component can be added to approximate the possible absorption bump. Overall, the representation for an extinction spectrum with an absorption bump can be expressed as:

\begin{equation}
\label{eq:bumps-1}
A(\lambda) =  c_1 + c_2x + c_3D(x, x_0, \gamma)
\end{equation}

, where $\lambda$ is the wavelength, and $x = \lambda^{-1}$. The component $D(x, x_0, \gamma)$ is a Drude profile, and it’s definition is:

\begin{equation}
\label{eq:bumps-2}
D(x, x_0, \gamma) = \frac{x^2}{(x^2 - x_0^2)^2 + x^2\gamma^2}
\end{equation}
, where $x_0$ and $\gamma$ is the peak position and full width at half maximum (FWHM) of the Drude profile, respectively. The strength can be measured by the area of the bump, i.e., $A_{bump} = \pi c_3/(2\gamma)$.  

\begin{figure}[tb]
\begin{center}
\includegraphics[width=0.7\linewidth]{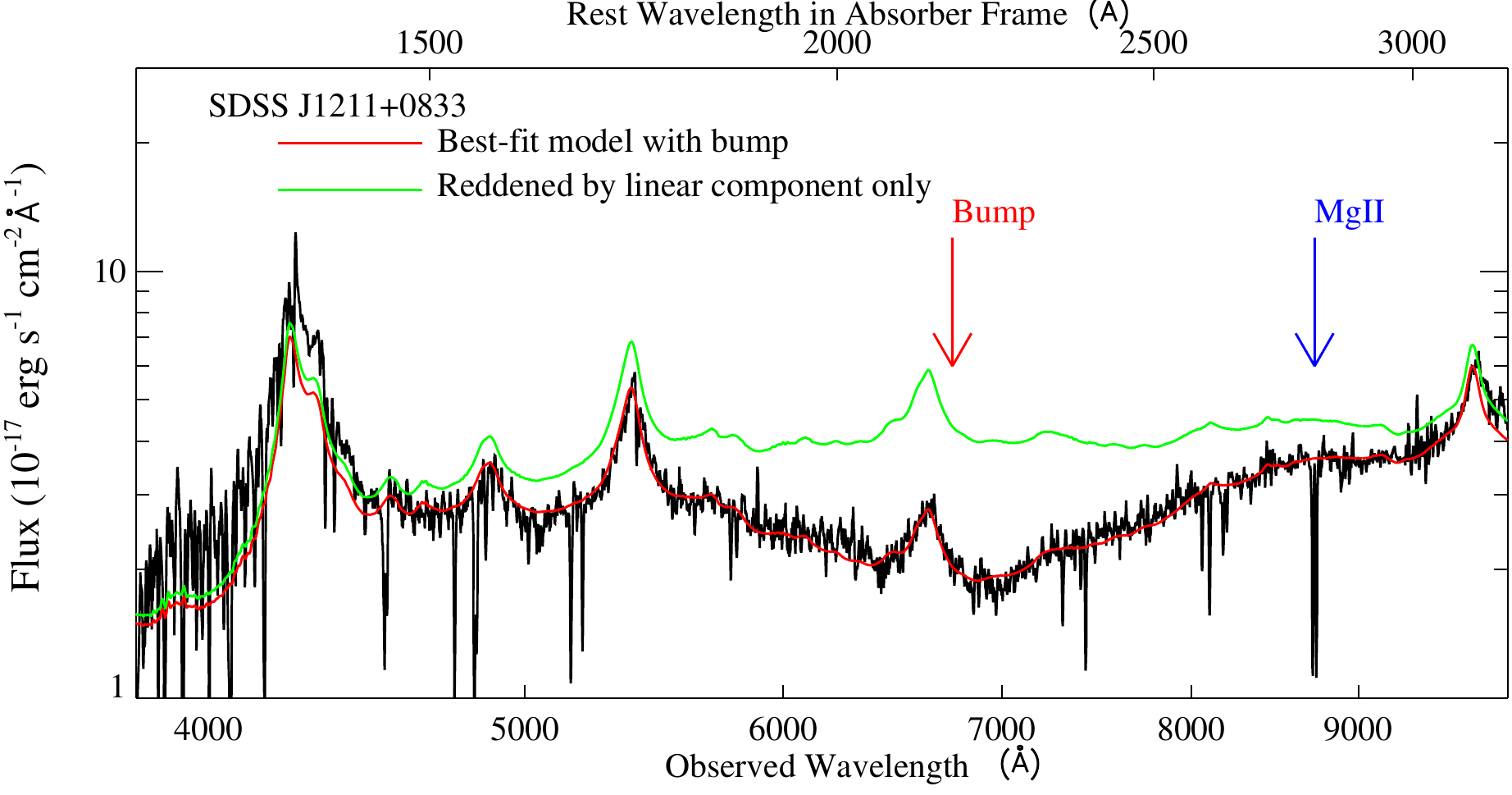}
\end{center}
\caption{Absorption Bump Example}
\label{fig:bump}
\end{figure}

The most popular method for discovering the 2175 \angstrom absorption bump is rule-based filtering after fitting. The main idea is to get the best fitted parameters for each observed quasar spectrum, following Equation~\ref{eq:bumps-1} and Equation~\ref{eq:bumps-2}. The whole procedure consists of three major steps. First, obtaining the median quasar composite spectrum. The basic assumption is that the quasars under similar condition will have similar observed spectrum. A median quasar composite spectrum is regarded as the base spectrum. It is created by combining the spectra from a set of observed quasars at their rest frame (e.g., \cite{fitzpatrick1990analysis, schneider2010sloan}). Second, curve fitting based on method such as Chi-Square Testing. The median quasar spectrum is first reddened to the quasar's emission redshift, and an absorption bump profile is convolved at the absorption redshift. The absorption redshift is obtained by referring to some unique absorption lines, such as the \ion{Mg}{2} absorption lines. The parameters in Equation~\ref{eq:bumps-1} and Equation~\ref{eq:bumps-2} are selected to minimize the fitting error. With the parameters, the last step is conducted by applying a set of rules to filter out low confidence candidates. One fitting example is shown in Figure~\ref{fig:bump}, where the black curve is the observed spectrum, the green curve is the reddened normal extinction spectrum continuum, and the red line is the continuum with absorption bump.

There are a couple of issues in this method. First, hard dependency on \ion{Mg}{2} absorption lines. The absorption redshift is obtained by referring to the \ion{Mg}{2} absorption lines. The \ion{Mg}{2} absorption lines are identified using some extra step, and is itself a research topic (\cite{quider2011pittsburgh, zhu2013jhu}). However, not all observations contain this absorption feature, due to different emission and absorption redshift ranges. Second, the error minimizing curve fitting method is not effective in some cases, e.g., broader Fe emission. This sometimes requires human identification. Third, the fitting procedure for each observation is time consuming. Multiple iterations of error minimizing steps are required to produce the best profile.

\section{Identifying the Absorption Bump with Deep Learning}
\label{sec:identify}


For the several aforementioned drawbacks of the traditional method, we propose a Deep Learning based method that could alleviate these issues and provide more flexible usage. Specifically, the proposed method exhibits the following merits.

\begin{itemize}
\item Large data set empowers accurate representation learning. The curve fitting step in the traditional method is repeatedly generating training samples. By carefully designing the generation, we are able to train an accurate model for the absorption bump detection.
\item No hard dependency on \ion{Mg}{2} absorption lines. When generating the samples, we are in control of selecting the emission and absorption redshifts. After the model is trained, no extra information is needed for new observations.
\item Easy to use and share. With a trained model, each prediction on newly added observations is just one feed forward process. Trained models could also be easily shared among researchers. Beyond trained models, simulation parameters and simulated data sets are all sharing-friendly. 
\item Easy to extend. Extensions of already trained models happen at two levels. First, additional training samples could be included to cover more cases. Second, multiple targets could also be added based on the same feature set. These extensions could be achieved conveniently by fine tuning of a trained model.
\end{itemize}

In order to validate the feasibility, we present our work on applying Deep Learning based method for the absorption bump detection in the following subsections. Specifically, we show the details of training data generation, and our experience on model selection and data transformation. 

\subsection{Raw Data Generation}
\label{sec:identify-raw}
We applied two methods to generate the data set. The first method follows the corresponding part from the curve fitting procedure. We refer to \cite{berk2001composite} and obtain the SDSS data release 7 composite quasar spectrum as the intrinsic spectrum. We then change the emission and absorption redshifts and add the absorption bump profile according to Equation~\ref{eq:bumps-2}. We choose a typical bump profile: $x_0=4.59$, $\gamma=1.0$ and $A_{bump} = 2.0$.

In order to increase the diversity and closely simulate the real-world observations, we apply the second method. We selected a subset of the spectra from the SDSS data release 7 according to the \ion{Mg}{2} catalog from \cite{zhu2013jhu}. We follow the fitting method described in \cite{jiang2010dusty, jiang2011toward}, and convolve them with the absorption bump profile same as the above one. These generated samples are extremely close to real-world cases. Note that we refer to existing \ion{Mg}{2} catalog to generate data. However, no dependency is required for predicting with new observations.

A total of 30K samples are generated with the two methods described above. Half of them contains the absorption bump and half does not. Each generated sample is stored in a file following the \textless wavelength, flux\textgreater\ pattern. There are around 4600 pairs in each file. We divide the data into two parts, 22K as the training data set and 8K as the testing data set. 

\subsection{Case One: Fully Connected Neural Network}
\label{sec:identify-fc}

\begin{table}[tbh]
\caption{Configurations and Test Results for Fully Connected Neural Network}
\label{tbl:conf}
\begin{center}
\begin{tabular}{ c c | c c | c c }
    \hline \hline
    \multicolumn{2}{c | }{One-Layer} & \multicolumn{2}{c | }{Two-Layer} & \multicolumn{2}{c}{Three-Layer} \\
    neurons & accuracy & neurons & accuracy & neurons & accuracy \\ 
    \hline
    10000   & 71.603\% & 800     & 96.658\%       & 800    & 95.540\%    \\
    1000    & 71.677\% & 600     & 96.485\%       & 600    & 95.888\%    \\
    600     & 71.727\% & 400     & \bf{96.758\%}  & 400    & 96.211\%    \\
    \hline \hline
    \multicolumn{2}{c | }{Four-Layer} & \multicolumn{2}{c | }{Five-Layer} \\
    neurons & accuracy & neurons & accuracy & & \\
    \hline
    800        & 95.826\%  & 800        & 96.298\% \\
    600        & 96.460\%  & 600        & 96.236\% \\
    400        & 95.603\%  & 400        & 96.124\% \\
    \hline \hline
\end{tabular}
\end{center}
\end{table}

The first straightforward choice for the model is the fully connected neural network. Its effectiveness in capturing the non-linearity relationships among the input features has been tested through the past decades. In order to feed the training data into the fully connected neural network, we extracted the flux value as a vector, and use this vector to train the model. We perform this data transformation based on the observation that the absorption bump is not related to the absolute wavelength value due to redshifts, but rather the neighboring flux value relationship. In addition, the wavelength interval between two consecutive wavelength-flux pair is relative stable in the data (both simulated and real-world). Due to the data collection mechanism, the number of pairs in one sample could be slightly different. We padded the samples with zeros to round up to a total of 4761 flux values. 

We changed the configurations of the fully connected neural network and vary the number of hidden layers and the number of neurons per layer. The output layer is softmax with two classes, with or without absorption bump. The initial learning rate is set to 0.01 and step decreasing police is applied. The testing results after converged training are shown in Table~\ref{tbl:conf}. The performance for one hidden layer is significantly worse than multiple-hidden-layer neural networks, due to its incapability to capture the non-linearity. The best performance is observed in a two-hidden-layer network with 400 neurons per layer. Its total number of trainable parameters is around 2 million.

\subsection{Case Two: Image and Convolutional Neural Network}
\label{sec:identify-image}

A quasar spectrum is plotted as the black curve in Figure~\ref{fig:bump} when inspected by researchers. This inspires us to transform raw data into images. We first plot all the training and testing samples. The coverage span for the wavelength is 8000 \angstrom and around 50 for the flux. The raw image drawn can be as large as hundreds of KB. The images are down scaled to 256*256 gray images and used for training and testing. Convolutional Neural Network (CNN) (\cite{lecun1995convolutional}) recently has proved its effectiveness in image related recognition and classification, due to its capability of parameter sharing and capturing localized patterns. We adapted two popular CNN based models, the AlexNet (\cite{krizhevsky2012imagenet}) and the GoogLeNet (\cite{szegedy2014going}). The results we get for modified AlexNet is 98.52\% and for GoogLeNet is 98.55\%. 

\begin{figure}[tbh]
\begin{minipage}{0.4\linewidth}1
    \subfloat[]{\includegraphics[width=\linewidth]{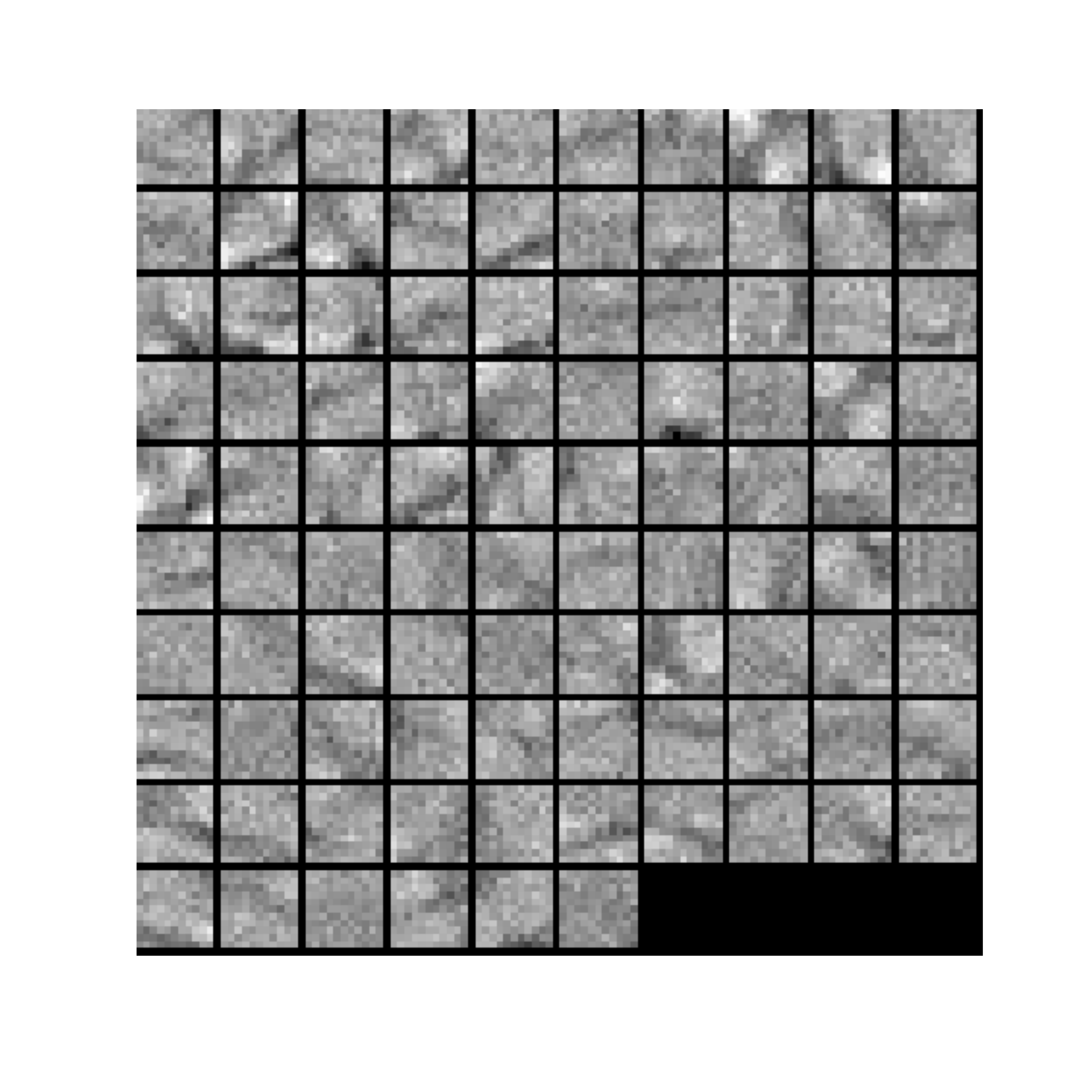}\label{fig:visual-filter}}
\end{minipage}
\begin{minipage}{0.25\linewidth}
    \center
    \subfloat[]{\includegraphics[width=0.6\linewidth]{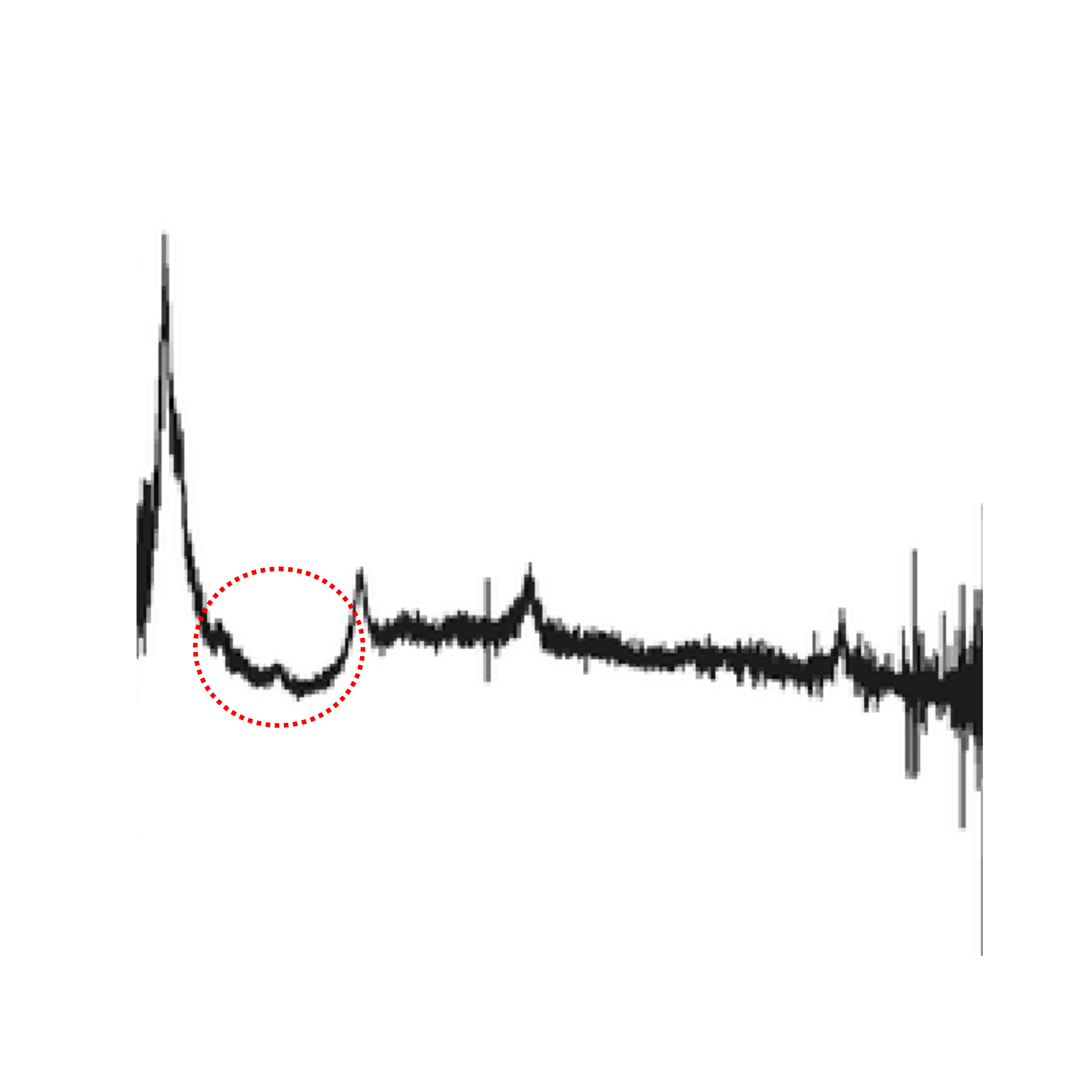}\label{fig:visual-left}}
    \vfill
    \subfloat[]{\includegraphics[width=\linewidth]{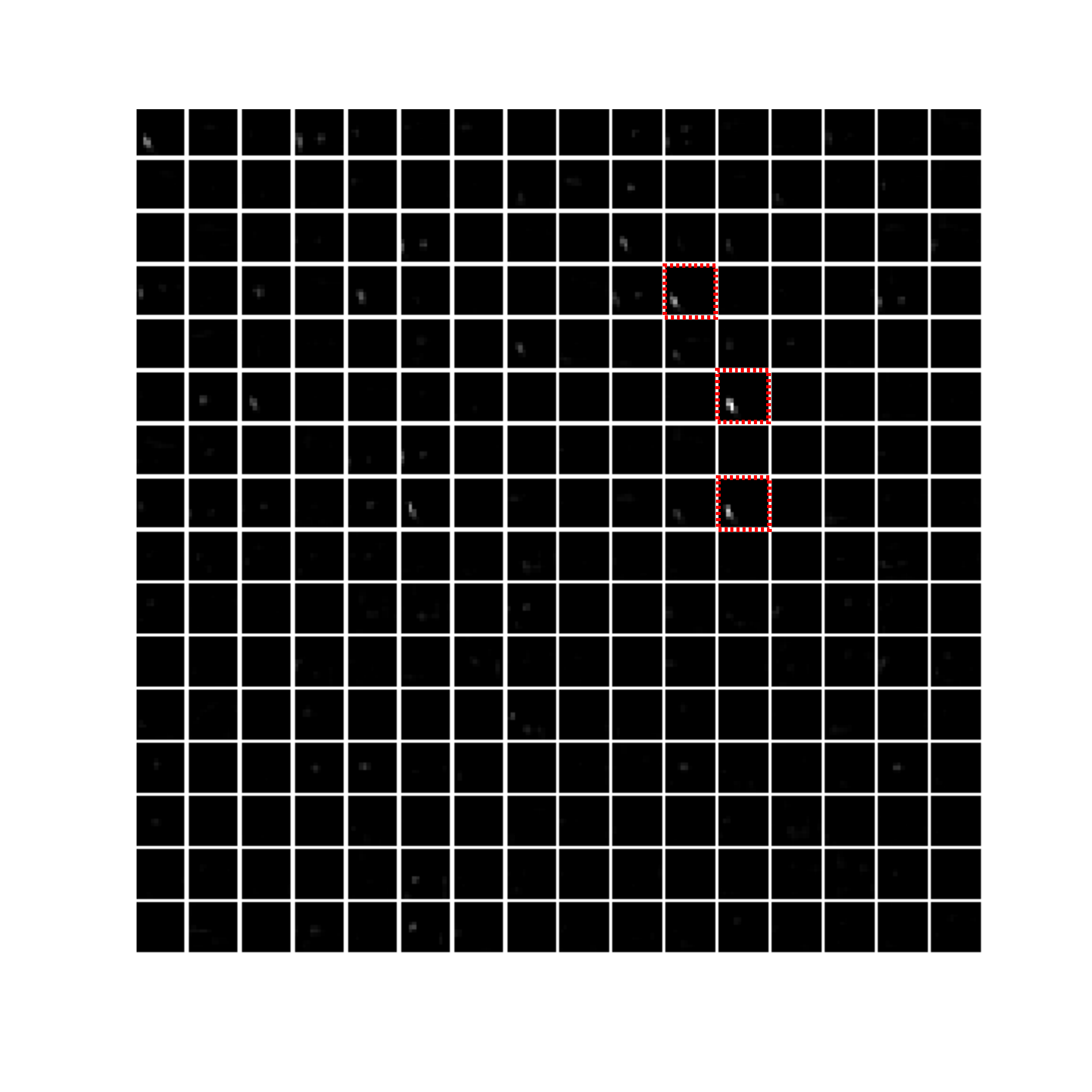}\label{fig:visual-fm-left}}
\end{minipage}
\begin{minipage}{0.25\linewidth}
\center
    \subfloat[]{\includegraphics[width=0.6\linewidth]{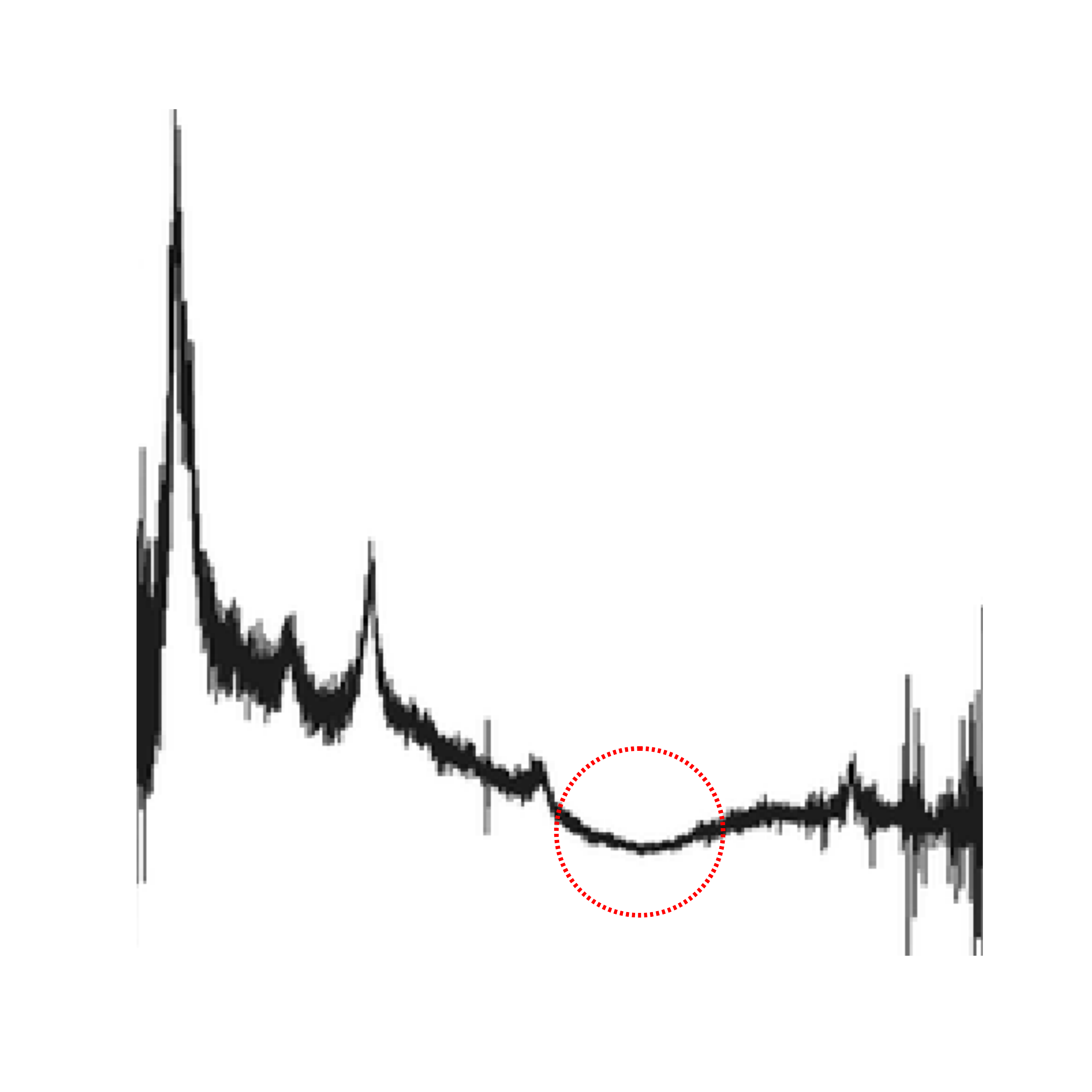}\label{fig:visual-right}}
    \vfill
    \subfloat[]{\includegraphics[width=\linewidth]{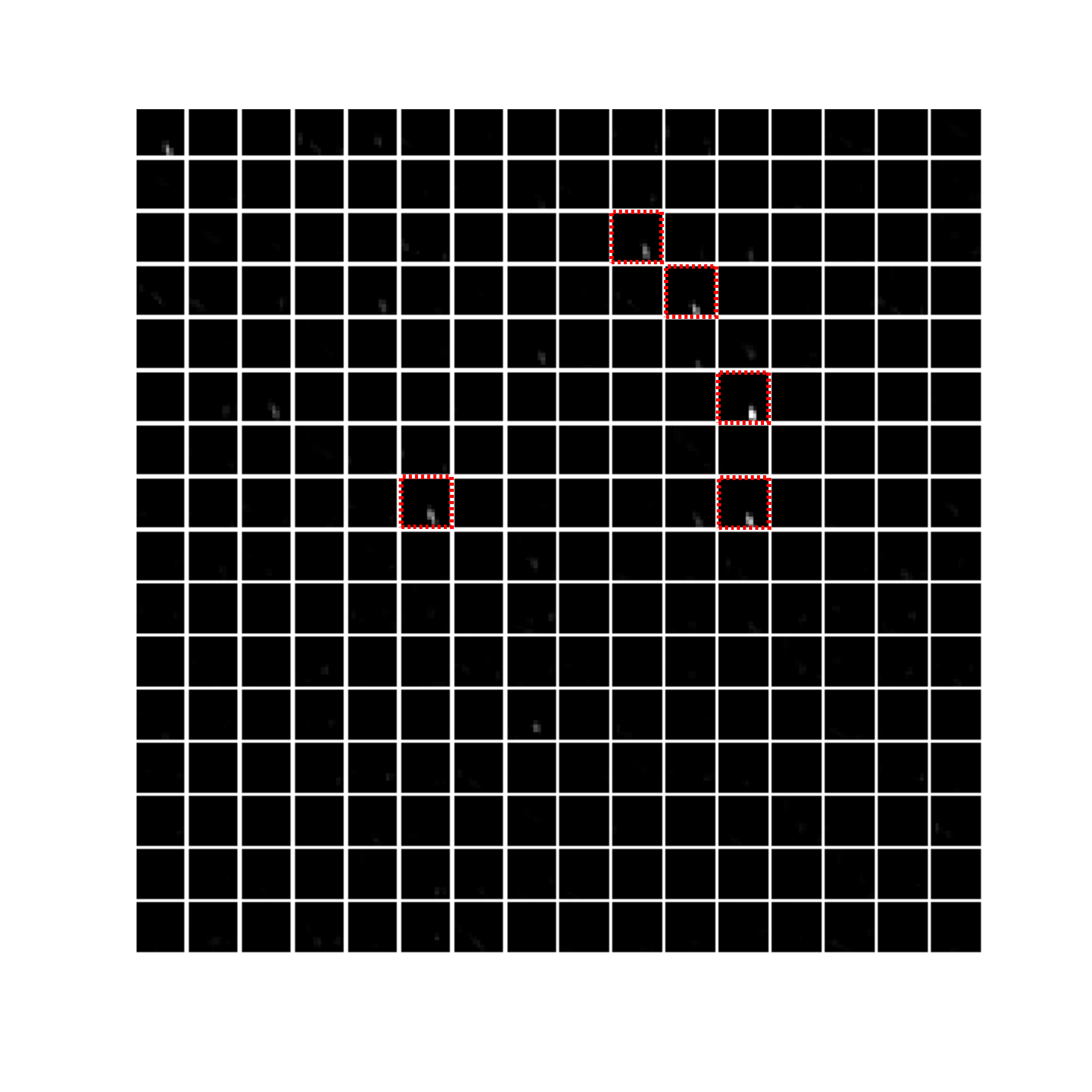}\label{fig:visual-fm-right}}
\end{minipage}
\caption{Visualization for AlexNet. \protect\subref{fig:visual-filter} shows the filters for the first convolutional layer. \protect\subref{fig:visual-left} and \protect\subref{fig:visual-right} plot two test images with absorption bump, marked with red dashed circle. \protect\subref{fig:visual-fm-left} and \protect\subref{fig:visual-fm-right} are the last convolutional layer feature maps for the two test images. The maps with effective activation are marked with red dashed rectangles. Better viewed in electronic version.}
\label{fig:visual}
\end{figure}

\begin{figure}[tb]
\begin{center}
\begin{tabular}{cc}
    \subfloat[Without Bump]{\includegraphics[width=0.3\linewidth]{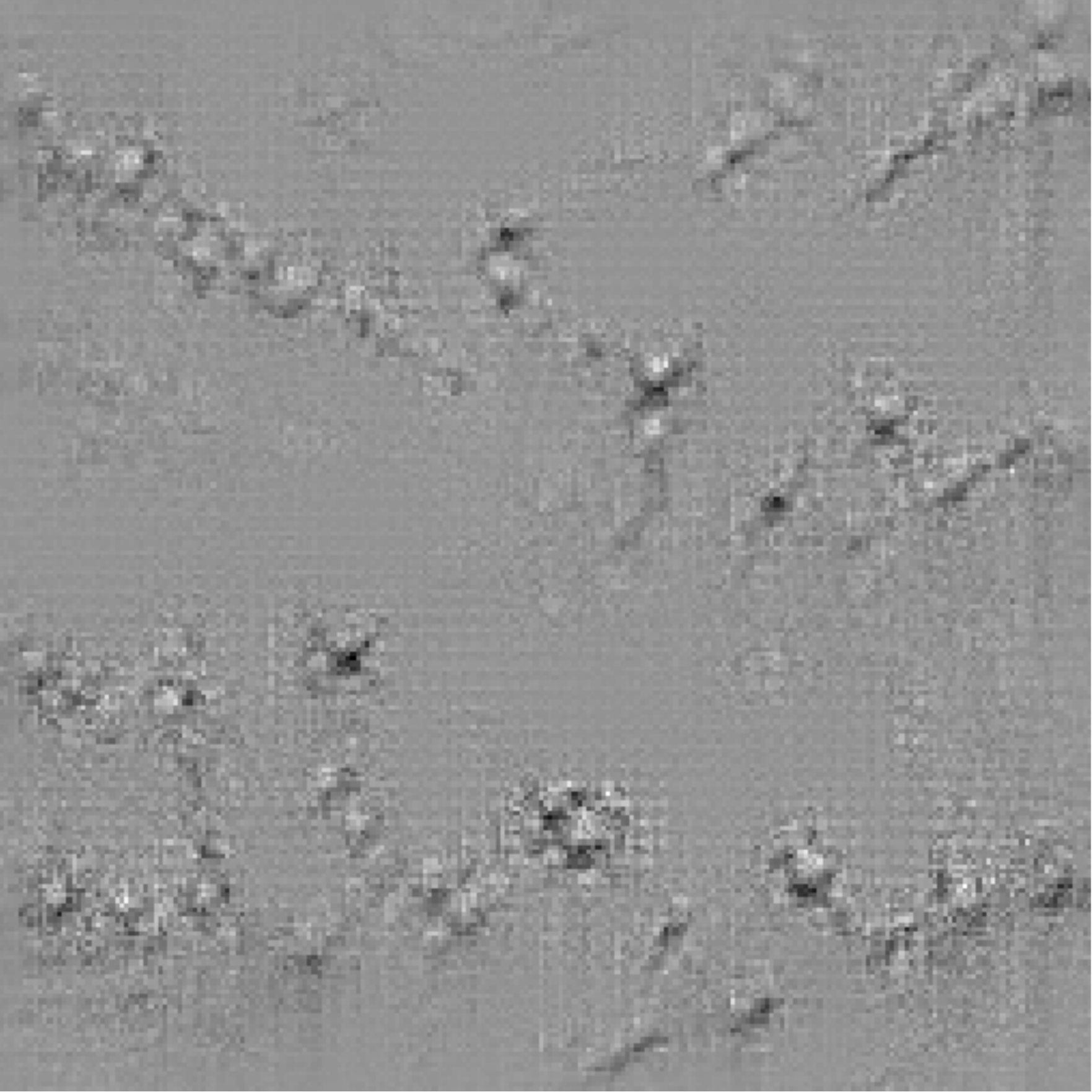}} & 
    \subfloat[With Bump]{\includegraphics[width=0.3\linewidth]{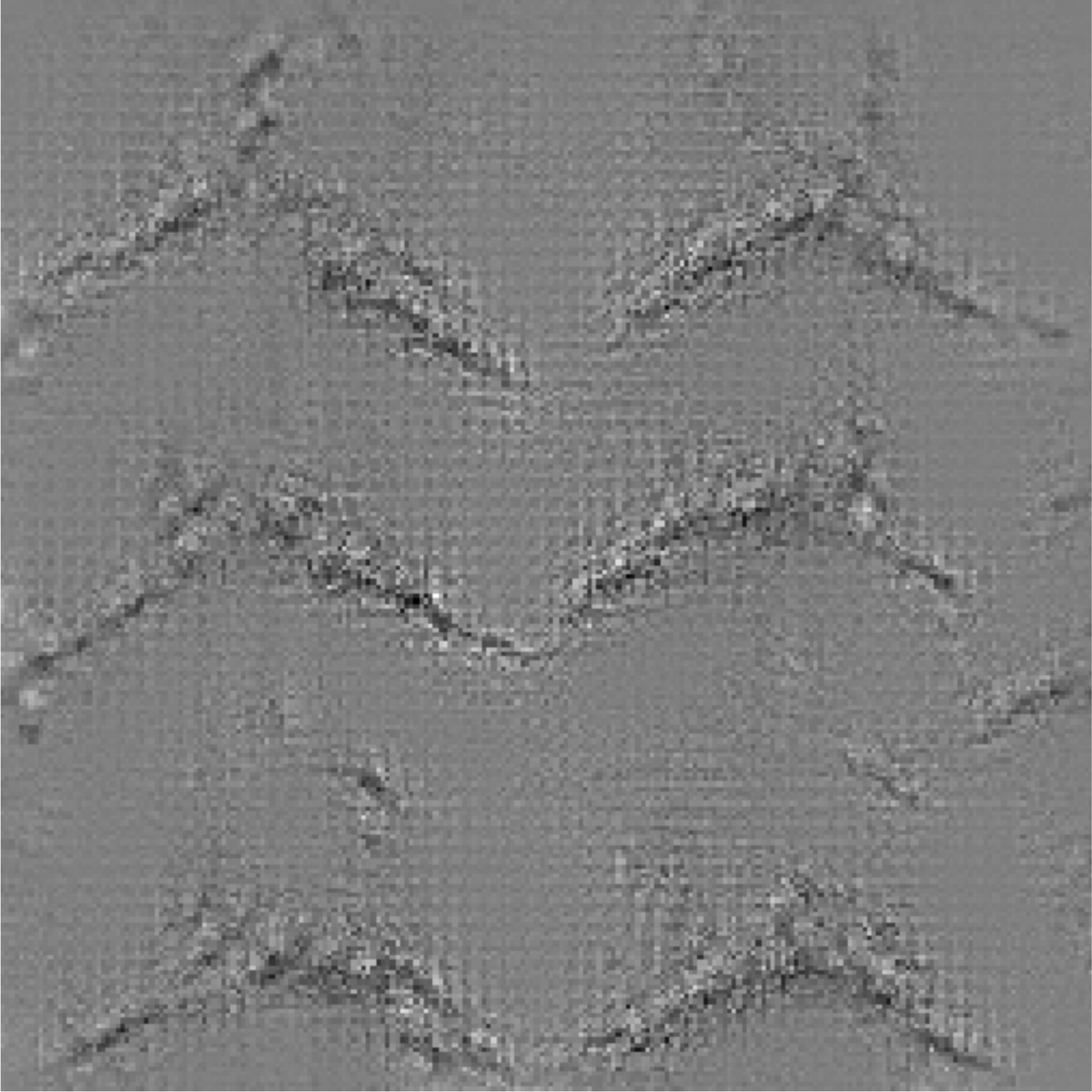}}
\end{tabular}
\end{center}
\caption{Reconstructed Input Image for with and without bump}
\label{fig:reconstructed}
\end{figure}

In order to understand the learned representation, we select the trained AlexNet model and visualize the first convolutional layer filters and also the last convolutional layer feature maps, as shown in Figure~\ref{fig:visual}. The first convolutional layer filters are typical edge detection filters. The effective activation area in the feature maps (Figure~\ref{fig:visual-fm-left} and \ref{fig:visual-fm-right}, marked with red dash rectangles) corresponds to the absorption bump locations in the test images. This proves that the representation is sensitive to the absorption bump. We also reconstructed the input images from a randomly generated image with respect to the given classes (with or without bump) using the gradient ascend method. The reconstructed images are shown in Figure\ref{fig:reconstructed}. The image for with bump clearly shows that the model is sensitive to the absorption bump in the input spectrum. 

\subsection{Case Three: Matrix and Convolutional Neural Network}
\label{sec:identify-cnn}

The 256*256 image plus CNN model improves the accuracy. However, the input image is too sparse considering the large white background. In order to condense the input as well as empower the CNN model, we perform another data transformation. This transformation is also based on the assumption used for fully connected neural network. Different from the previous one, where the input is a 1D vector, we pad and fold the vector into a 69*69 matrix. These folded matrices are fed into the CNN models. This transformation makes it possible for filters in the convolutional layer to capture the localized information with neighboring flux values. We use different configurations of the CNN and post the best results we find in Table~\ref{tbl:cnn69}. We fix the CNN models with two fully connected layers after the convolutional layers, and alter the number of convolutional layers, the number of filters in each convolutional layer, and the kernel size. The best model we get achieves 99.404\% test accuracy, and is a model with 4 convolutional layers, each with 50 filters, and the filter size is 5*5, 3*3, 3*3, 3*3 for the convolutional layers, respectively. The trainable parameters is around 900,000, instead of several or tens of millions in the previous two cases.

\begin{table}[tbh]
\caption{Configurations and results for CNN on 69*69 Matrix}
\label{tbl:cnn69}
\begin{center}
\begin{tabular}{ l l }
    \begin{tabular} {c c}
    \hline \hline
    \multicolumn{2}{c}{\textbf{Configurations}} \\
    \hline
    Conv Layers & 2, 3, 4 \\
    Filters     & 40, 50, 60, 70 \\
    Kernel Size & 7*7, 5*5, 3*3 \\
    \hline \hline
    \end{tabular}
    &
    \begin{tabular} {c c}
    \hline \hline
    \multicolumn{2}{c}{\textbf{Best Results}} \\
    \hline
    CNN2 & 98.696\% \\
    CNN3 & 99.155\% \\
    CNN4 & 99.404\% \\
    \hline \hline
    \end{tabular}
\end{tabular}
\end{center}
\end{table}

\subsection{Discussion}
\label{sec:identify-discuss}

\begin{figure}[th]
\centering
    \begin{minipage}[h]{.55\textwidth}
        \centering
        \includegraphics[width=\textwidth]{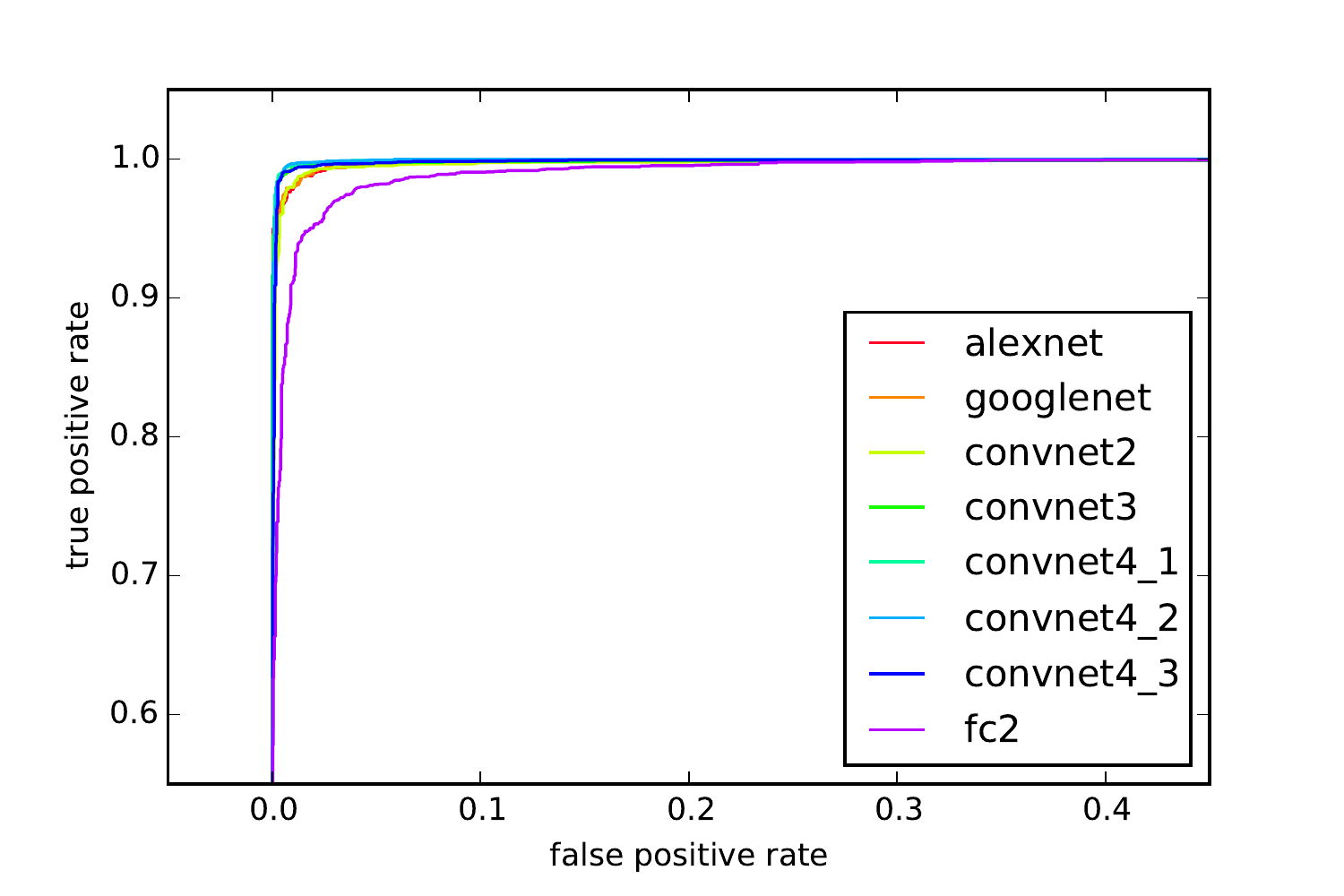}
        \caption{ROC curves for various models}
        \label{fig:roc_figs}
    \end{minipage}
    \hspace{0pt}
    \begin{minipage}[h]{.4\textwidth}
        \centering
        \captionof{table}{AUC of the ROC curves}
        \label{tbl:auc}
        \begin{tabular} {c c}
        \hline \hline
        \textbf{Type} & \textbf{AUC} \\
        \hline
        AlexNet    & 0.99915 \\
        GoogLeNet  & 0.99898 \\
        ConvNet2   & 0.99802 \\
        ConvNet3   & 0.99873 \\
        ConvNet4-1 & 0.99943 \\
        ConvNet4-2 & 0.99947 \\
        ConvNet4-3 & 0.99875 \\
        FC2        & 0.99455 \\
        \hline \hline
        \end{tabular}
    \end{minipage}
\end{figure}

In previous sections, we present three kinds of models and the corresponding data transformations. We select some of the trained models and plot their ROC curves, shown in Figure~\ref{fig:roc_figs}, and the corresponding AUCs are given in Table~\ref{tbl:auc}. The ROC curves for the selected models are consistent with the previously reported testing accuracy. For this absorption bump detection application, we could improve the sensitivity of the trained models by slightly decreasing the decision threshold. This could result in slightly more false positive signals, but is more confident in capturing the potential absorption bump events we are interested in.

Based on our experience with absorption bump detection application, there are a couple of things worth notice for data preparation and model selection. When generating raw data, in order to eliminate the bias in the trained model, carefully designed criteria should be followed to cover the cases as much as possible. Another necessary step before feeding the data into training is shuffling. The raw data are generated in a systematical manner. Using non-shuffled data, localized characteristics in the training batches could impede the convergence of the model. It could also cause diverged performance on training and testing data set. We have experienced these problems at early stage. The accuracy for the best model we get from non-shuffled data is as low as 88\%. 


From the presented details, we believe that Deep Learning based method for absorption bump detection is effective. With a trained model, \ion{Mg}{2} information is not required to filter new observations for candidates. The specifications for data generation and model configuration, and the trained model are easy to share and reuse. We observe from our experience that the data generation could also be modeled into a MapReduce-like programming model, significantly relieving users from heavy programming, but still benefiting from the power of distributed computing. In the following section, we present our generalization and current work on providing such a Deep Learning based framework for similar problems.

\section{Generalization: Boosting Science Discovery with Deep Learning}
\label{sec:gen}


\subsection{Science Discovery Problem}
\label{sec:gen-problem}
Similar as the absorption bump detection, a category of problems in other science field involves the detection of a certain phenomenon among the background events. We call them the Science Discovery Problem. These discoveries are building blocks for further scientific research. Scientists rely on them to propose ideas, validate hypotheses, and prove theories. A typical science discovery problem consists of two main components, the feature set, $\mathbb{F} = \{f_i|0<i<=N\}$, and the relation among the features, $\mathbb{R}$. A science discovery problem can be expressed as:
\begin{equation}
\label{eq:problem}
\mathbf{D} = \{\mathbf{F} | \mathbf{F} \sim \mathbf{R}_{target}(P, C), \mathbf{R}_{target} \subset \mathbb{R}\}
\end{equation}

$\mathbf{D}$ is the targeted discoveries, $P$ and $C$ are parameters and constants. $\mathbf{F}$ is one event in $\mathbb{F}$. $\mathbf{R}_{target}$ defines the relations among the features that depict the targeted phenomenon. $\mathbf{R}_{target}$ and its complementary set $\mathbf{R}_{target}^{\textbf{c}}$ compose $\mathbb{R}$.

With the $\mathbb{F}$, $\mathbb{R}$, $\mathbf{R}_{target}$, scientists aim to pick the discoveries out of the observations. The mainstream method to perform the detection is rule-based filtering. The filtering rules are constructed upon raw features or extracted high level features. High level feature extraction can be done by numerically combining raw features, or by some fitting method, similar as the absorption bump application. While scientists are concerned with the accuracy of the approximation theories and continue improving their effectiveness, there are some issues with such methods.


\textbf{Feature Engineering.} Feature engineering is a time-consuming procedure. It also depends on a lot of experience and numerous trial-and-error experiments. The complex and mysterious nature of the science discovery problems also increase the difficulty of extracting meaningful high level features to effectively capture the target event characteristics. Even with a carefully developed feature extraction technique with excellent theoretical explanation, real-world data could also offset the expected effectiveness considering the existence of various noise and uncertainty during the data collection phase. 

\textbf{Extra Dependency.} In some science discovery cases, in order to get a better approximated representation for the targeted event, extra effort is required to obtain additional information. For example, the approximated absorption bump representation requires emission and absorption redshifts. While the emission redshift is included in the public data set, the absorption redshift is calculated based on \ion{Mg}{2} absorption lines. Similar requirements occur in other science fields. Two potential drawbacks in such scenarios impede the science exploration. First, these extra dependencies incur considerable amount of time and efforts. Second, they may not co-exist with the target discovery, thus, restricting the exploration scope due to incomplete data sets. 

\textbf{Sharing and Collaboration.} Information exchange and collaboration are two significant factors to speed up the science discoveries. The theories and findings are well shared by researchers, but heavy data processing are required to generate results with existing methods for each new observation. Even though distributed computing technologies are pervasive and performance-ascendant, cross-discipline knowledge requirement impedes its wide adoption.

\subsection{Deep Learning based Method}
\label{sec:gen-dl}


Inspired by the advance and practical effectiveness of Deep Learning, we propose a data-driven, simulation-based method for science discovery problems. The proposed method is rooted in three observations. First, Deep Learning has proven itself with practical success on classification and recognition problems. Second, Deep Learning is more powerful when the relations among features are complicated. This is often the case in science discovery problems. Finally, Deep Learning requires a large amount of training data to be effective. Despite of the fact that real-world data might not be adequate and they are sometimes biased (i.e., the targeted discoveries are normally rare compared with the background events), we observe that it is relatively easy to simulate enough eligible training samples with desired information embedded in. The control over the training data sets ensures us to accurately capture the targeted phenomena. This both solves the biased training issue and satisfies Deep Learning requirement for large-scale training samples. 

\begin{figure}[tbh]
\begin{center}
\includegraphics[width=0.7\linewidth]{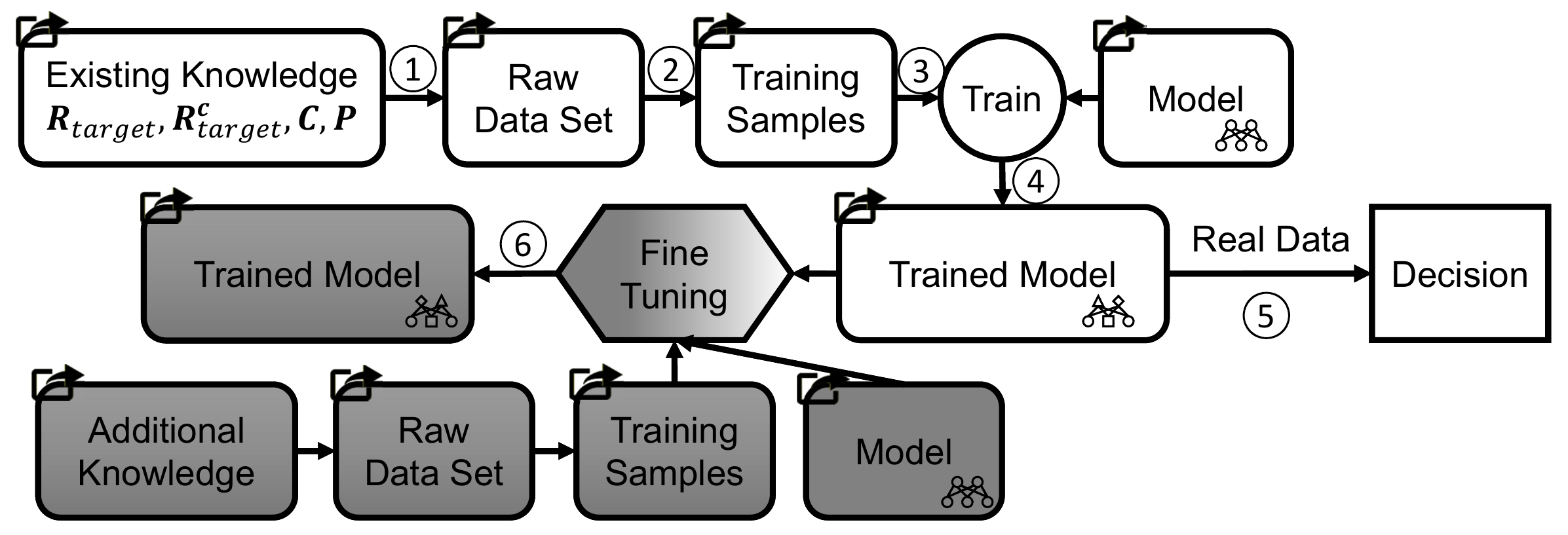}
\end{center}
\caption{Workflow for DeepDis. The mark on the top left corner means able-to-share.}
\label{fig:workflow}
\end{figure}

The workflow of the proposed method is shown in Figure~\ref{fig:workflow}. There are three main steps: data preparation, model training, and model service. During data preparation, two operations are performed: data generation (1) and data transformation (2). The approximation theory $\mathbf{R}_{target}$ and $\mathbf{R}_{target}^{\textbf{c}}$, the constant variables $C$, and a set of range parameters $P$ are used to generate enough labeled raw data. The raw data can be represented by a series of key-value pairs. The data transformation phase maps the raw data sets into desired formats for training, e.g., images or multi-dimensional matrix. The transformed data sets are used in model training step to train a pre-defined model (3). After the converge criteria are satisfied, the trained model will be published and ready for use (4). In model service step, the trained model is used to give decisions for real-world data (5). The real-world data set go through similar transformations as the simulated raw data sets. 
In science discovery problems, various phenomena could be detected upon the same feature set. This inspires us to introduce the fine tuning and transfer learning techniques into the workflow (6). Specifically, additional data could be generated to further tune an already trained model. By substituting some layers in the previous model and apply the fine tuning using extra data sets, a new model could be trained to identify the new phenomenon. 

\begin{figure}[tbh]
\begin{center}
\includegraphics[width=0.6\linewidth]{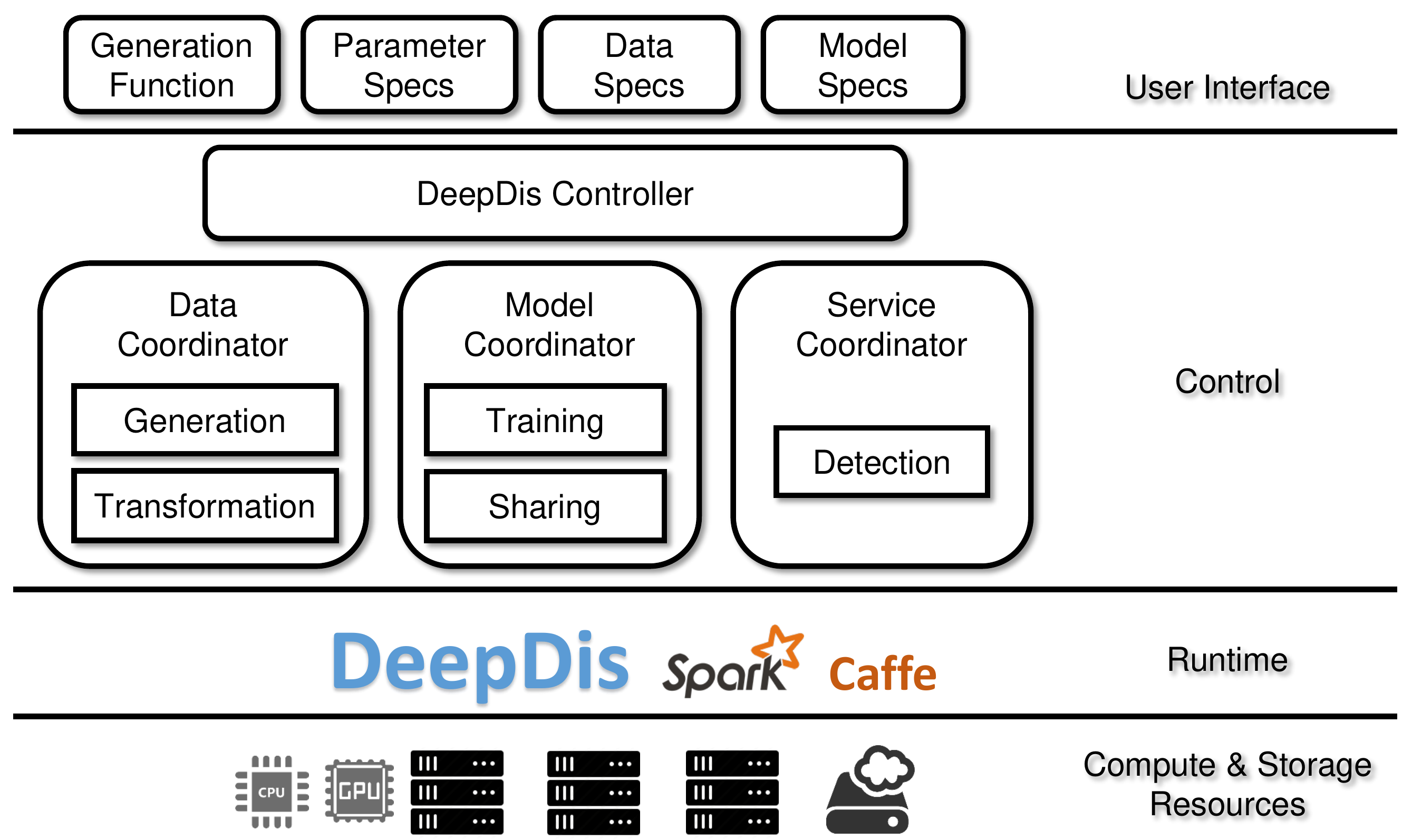}
\end{center}
\caption{Architecture Overview for DeepDis}
\label{fig:arch}
\end{figure}

The architecture of the proposed DeepDis system is shown in Figure~\ref{fig:arch}. There are four layers in DeepDis, namely user interface layer, control layer, runtime layer, and compute and storage resource layer. From the beginning of the design, several main considerations are planted in the gene of DeepDis. The first one is flexibility and easy-to-use. The second one is harnessing the power of various computing and storage resources. The third one is sharing. We show how we emphasize on these aspects in the following descriptions.

\textbf{User Interface Layer}. Two kinds of information are required from the users. The first part is the data generation function and parameters. Users are requested to implement their own generation functions. The function will take a dictionary of parameters, including constants and range parameters, and output all possible key-value pairs constrained by the ranges. It resembles the flat map function in many programming languages. Data specifications also contains instructions on how to transform raw data sets into ready-to-process data format. The second part is the model configuration. The specifications are passed to DeepDis by json/prototxt. Only these configurations are required from the users. This makes DeepDis easy-to-use even for scientists with less knowledge about distributed computing or Deep Learning.

\textbf{Control Layer}. This layer is responsible for understanding users requests and translate them into computing tasks. Specifically, DeepDis has a controller and three coordinators, Data Coordinator, Model Coordinator and Service Coordinator, to fulfill this. The controller divides the requests into several inter-dependent tasks and use DAG to schedule. The coordinators are responsible for the designated tasks: preparing, dividing, and distributing them to the underlying runtime. This layer also opens pluggable and extensible interfaces for advanced users to provide new data format, transformation functions, etc.

\textbf{Runtime Layer} and \textbf{Compute and Storage Resources Layer}. DeepDis integrates state-of-the-art distributed in-memory computing framework, Spark (\cite{zaharia2012resilient}), and the Caffe (\cite{jia2014caffe}) Deep Learning toolbox to efficiently execute the data generation and model training tasks. Spark will be extended to incorporate adaptations from the control layer and to collaborate with Caffe for distributed model training. For the Compute and Storage Resources Layer, different types of resources will be utilized to achieve efficient data processing, model training, and sharing.

\section{Conclusion}
\label{sec:conclusion}

In this paper, we presented a Deep Learning based method for detecting the absorption bump. We discussed in detail about how we generated and transformed the training data according to the selected models. For different models, we presented the results. In order to get a sense of what the models have learned, we chose the CNN based AlexNet model and provided the visualizations of filters, feature maps and reconstructed maximum activation input images. The resulted images prove that the model is identifying the absorption bump, rather than any other background noise. With the success in applying Deep Learning based method for the absorption bump application, we generalized the methodology to the broad science discovery problems, where Deep Learning models can be effectively trained upon sufficient amount of data. We also showed our ongoing work in building a specialized system, DeepDis, to support the proposed method. Distributed data processing techniques are used to automatically handle data generation, data transformation, model training and sharing. DeepDis is designed to provide science discovery as a service for researchers without too much knowledge about distributed computing and Deep Learning. With DeepDis, we hope to boost the science discovery process.


\bibliography{bumps-iclr2016-bibfile}
\bibliographystyle{iclr2016_conference}

\end{document}